\documentclass[letterpaper, 10 pt, conference]{ieeeconf}  

\IEEEoverridecommandlockouts                              

\usepackage{graphics} 
\usepackage{epsfig} 
\usepackage{subfig}
\usepackage{amsmath} 
\usepackage{amssymb}  
\usepackage{booktabs} 
\usepackage{cite}

\title{\LARGE \bf
Learning and Generalizing Motion Primitives from Driving Data for Path-Tracking Applications
}

\author{Boyang Wang$^{1,2}$,~\IEEEmembership{Student Member,~IEEE}, Zirui Li$^{1}$, Jianwei Gong$^{1}$,~\IEEEmembership{Member,~IEEE}, \\Yidi Liu$^{1,3}$, Huiyan Chen$^{1}$, Chao Lu$^{1}$
\thanks{*This work was supported by the National Natural Science Foundation of China (No.91420203 and No.61703041)}
\thanks{$^{1}$ All the authors are with the School of Mechanical Engineering, Beijing Institute of Technology, Beijing, China, 100081
        {\tt\small gongjianwei@bit.edu.cn}}%
\thanks{$^{2}$ Boyang Wang is also with the Interactive Digital Human group of CNRS-UM LIRMM, UMR5506, Montpellier, France, 34095.
        {\tt\small wbythink@hotmail.com}}%
\thanks{$^{3}$ Yidi Liu is also with the Advanced Technology Development Department of SAIC Motor Corporation Limited, Shanghai, China, 201804.
        {\tt\small liuyidi\_rosamond@163.com}}%
}

\begin{document}

\maketitle
\thispagestyle{empty}
\pagestyle{empty}

\begin{abstract}

Considering the driving habits which are learned from the naturalistic driving data in the path-tracking system can significantly improve the acceptance of intelligent vehicles. Therefore, the goal of this paper is to generate the prediction results of lateral commands with confidence regions according to the reference based on the learned motion primitives. We present a two-level structure for learning and generalizing motion primitives through demonstrations. The lower-level motion primitives are generated under the path segmentation and clustering layer in the upper-level. The Gaussian Mixture Model (GMM) is utilized to represent the primitives and Gaussian Mixture Regression (GMR) is selected to generalize the motion primitives. We show how the upper-level can help to improve the prediction accuracy and evaluate the influence of different time scales and the number of Gaussian components. The model is trained and validated by using the driving data collected from the Beijing Institute of Technology (BIT) intelligent vehicle platform. Experiment results show that the proposed method can extract the motion primitives from the driving data and predict the future lateral control commands with high accuracy. 

\end{abstract}

\section{Introduction}

Intelligent vehicles are expected to be integrated into our daily life. This requires the vehicles to perform human-like actions which can improve the adaptability for human driving habit and increase the potential for acceptance of intelligent driving system\cite{wilhelem2017energy,schnelle2017driver,schnelle2017personalizable}. Learning by imitation or learning from demonstrations is an effective way that enables the vehicles to learn from observing motions and to generalize the learned motions to new situations\cite{argall2009survey}. Approaches which are utilized to represent motions have been developed as a typical framework for imitation learning. The observed motions are often represented by abstract forms and these abstract forms are called motion primitives (MPs)\cite{ijspeert2002movement}\cite{paraschos2013probabilistic}. MPs have been well-established and used to solve many complex tasks in robotics, such as unscrewing a light bulb\cite{manschitz2014learning}, striking table tennis\cite{mulling2013learning} and ball throwing\cite{ude2010task}. The purpose of using MPs is to enable the decomposition of complex tasks into elemental movements and develop more efficient solution algorithms.

\par Based on the same concept, Bender et al.\cite{bender2015unsupervised} developed a two-layer unsupervised approach for inferring driver behavior from driving data. The driving data include the pose and dynamic state of the vehicle are decomposed into linear segments automatically in the first layer. And the driving behavior such as turning, braking, accelerating and coasting are inferred from the driving data within each segment in the second layer. Taniguchi et al.\cite{taniguchi2015unsupervised} proposed an unsupervised hierarchical model to predict driving behavior using driving data which consist of steering angle, brake pressure and accelerator pedal position. A sequence of driving words is generated from the driving data, each consisting of several driving letters. And the driving letters which represent one typical driving behavior are then inferred within each driving word\cite{taniguchi2016sequence}. 
\begin{figure*}[t]
\centering
\subfloat[A total of 81 hours of driving data are collected by the platform. The steering-wheel angle which collected from CAN bus network and the path data which are gathered from the integrated navigation unit are used to train the motion primitives in this paper.]{
\begin{minipage}[t]{0.45\textwidth}
\centering
\includegraphics[scale=0.32]{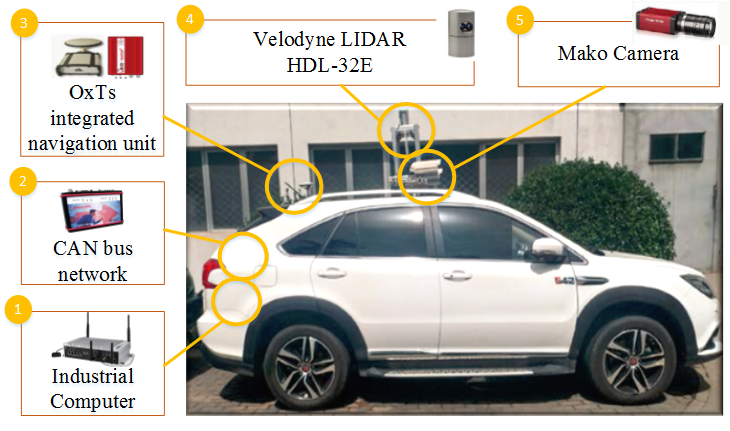}
\end{minipage}
\label{fig１(a)}
}
\hspace{10pt}
\subfloat[Sampled data of one typical demonstration. The background color indicates the identified path type.The selected clustering features include the time duration, average and maximum course deviation and average velocity of the segmented path primitive.]{
\begin{minipage}[t]{0.45\textwidth}
\centering
\includegraphics[scale=0.38]{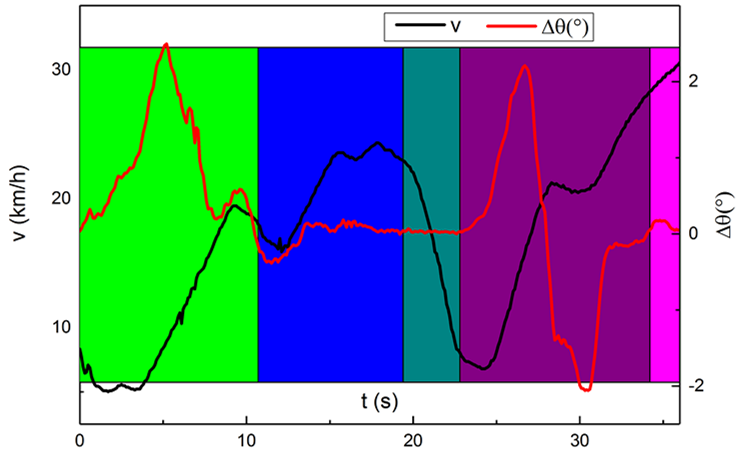}
\end{minipage}
\label{fig１(b)}
}

\subfloat[The path types are identified by using segmentation and clustering method in the upper-level. In the lower-level, the motion primitives are trained based on the regrouped data.]{
\begin{minipage}[t]{0.45\textwidth}
\centering
\includegraphics[scale=0.25]{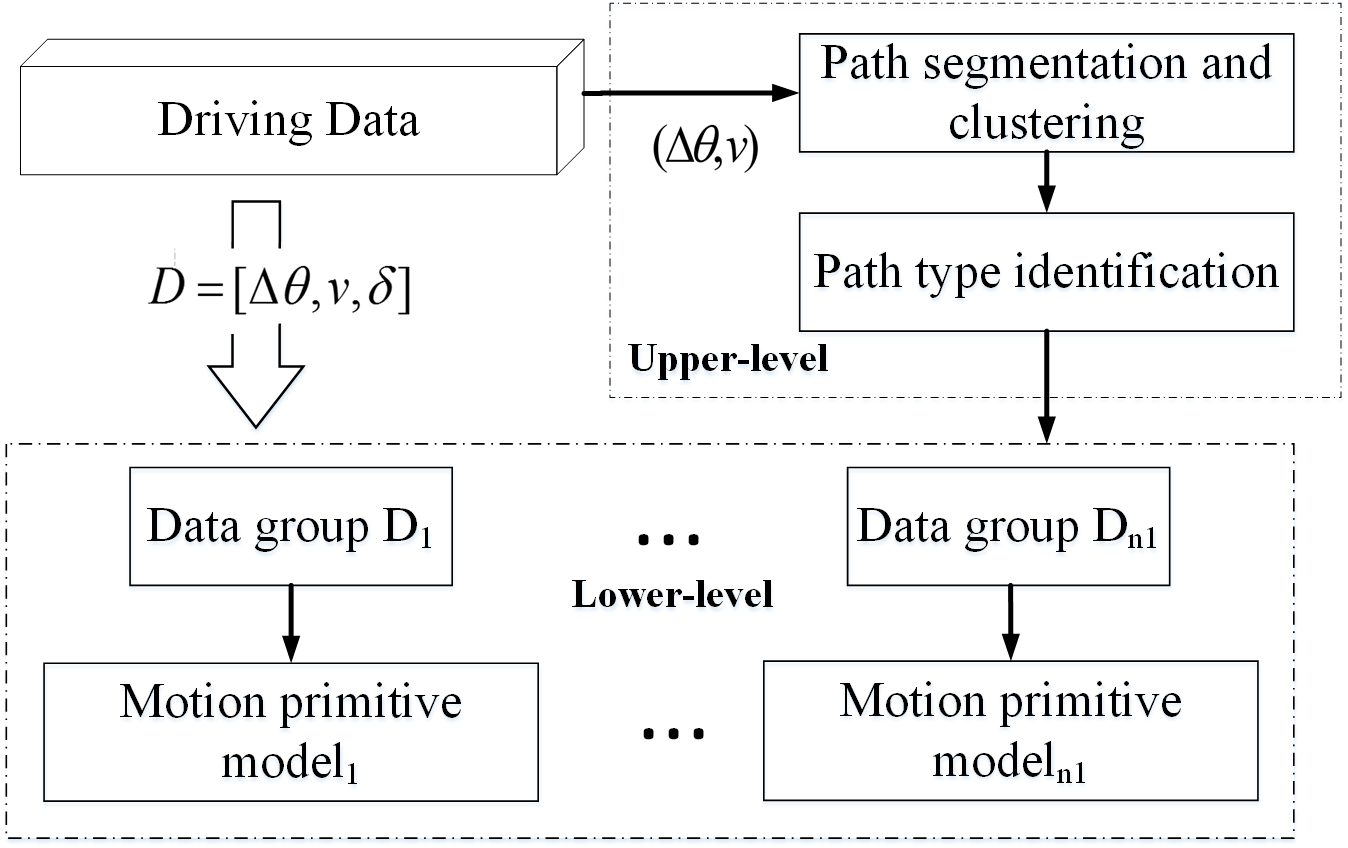}
\end{minipage}
\label{fig１(c)}
}
\hspace{10pt}
\subfloat[The motion primitives take into account the path and operation commands of previous, current and future. The steering wheel angles are predicted with uncertainty description by generalizing the learned motion primitives. ]{
\begin{minipage}[t]{0.45\textwidth}
\centering
\includegraphics[scale=0.3]{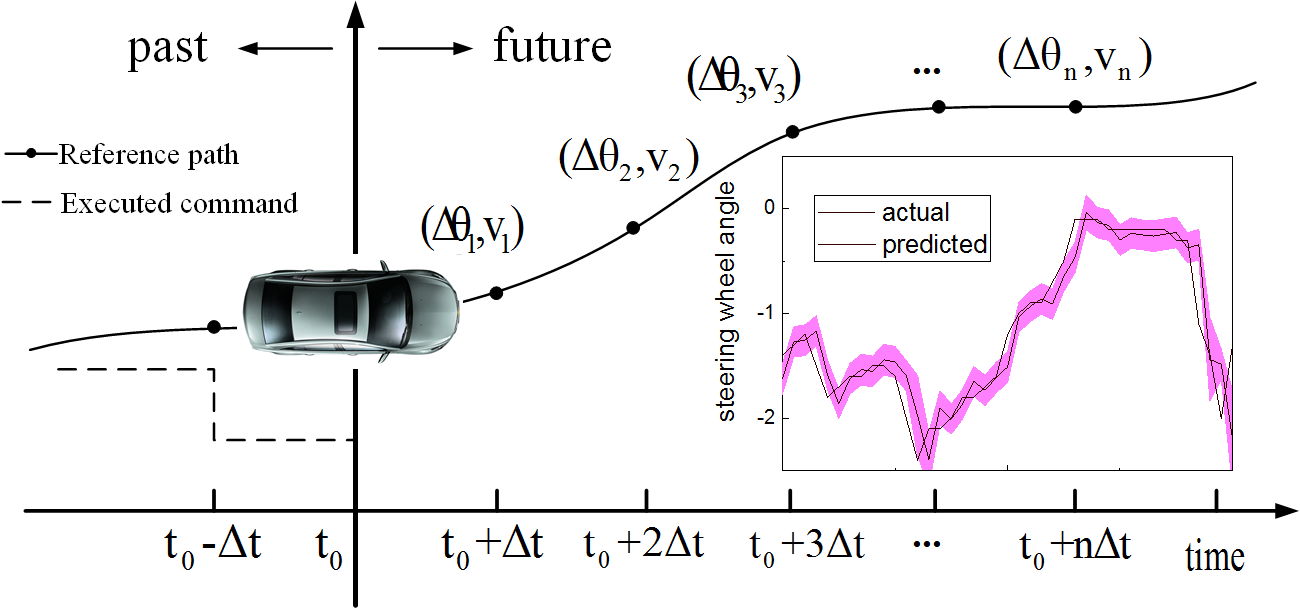}
\end{minipage}
\label{fig１(d)}
}
\caption{The overall flow of our framework. First, the driving data are collected by the BIT intelligent vehicle platform (a). Then, the path types are identified in the upper-level (b,c). According to the results of the upper-level, the motion primitives are learned and generalized in the lower-level of our framework (c,d)}
\label{fig１}
\end{figure*}
\par Although the MPs are extracted from driving data, the purpose of the above-mentioned approaches are to predict discrete results of driving behaviors, not continuous driver operation behaviors. Inspired by the work of Taniguchi, Hamada et al.\cite{hamada2016modeling} propose to apply the beta process autoregressive hidden Markov model (BP-AR-HMM) with dynamical systems to predict general driving operation behaviors, not the sequence of symbols. Flad et al.\cite{flad2017cooperative} identified a model which describe the driver steering behavior as a sequence of MPs. Besides, each MPs describe the resulting steering angle sequence under a specific configuration of the driver's neuromuscular system. However, the application of these methods is limited in the field of driving assistance systems. Their missions are to infer the future driving behavior with the past driving data and parameterized driving situations, not to predict the driving behavior with the desired path informations which are generated by the motion planner or navigation system.

\par Predictive control is often applied to improve the efficiency and performance of the path-tracking controller. One crucial aspect is to predict future vehicle motions considering the reference path. Ostafew et al.\cite{ostafew2016learning} established the motion prediction model with a priori vehicle model and a learned disturbance model. Disturbances are modeled as a Gaussian Process and updated based on the experience from collected data during previous trials. As a result, the proposed method is limited to reduce trajectory-specific path-tracking errors. Funfgeld et al.\cite{funfgeld2017stochastic} proposed an approach which can predict the vehicle motion considering the available road data and uncertain driver behavior. The road data are given as a tree-like structure of road segments and the driver behavior are defined by the maximum tolerated vehicle states of each road segment.

\par Differing from previous research focusing on the past driving data and parameterized driving situations\cite{bender2015unsupervised,taniguchi2015unsupervised,taniguchi2016sequence,hamada2016modeling,flad2017cooperative}, the future path reference is also taken into consideration in our MPs. Besides, neither specific trajectory\cite{ostafew2016learning} nor segmented road data\cite{funfgeld2017stochastic} has been used to establish the MPs. The general path data that are collected from our data collection system is applied as training data for our MPs. In order to predict the steering wheel angle based on personalized driver behavior, a two-level MPs learning and generalizing structure has been proposed, in which the lower-level motion primitives have to be extracted and generalized under the upper-level path segmentation and cluster layer as shown in Fig. \ref{fig１}. The main task of our work will be explained further in Section \uppercase\expandafter{\romannumeral2}. Furthermore, the influence of different time scales and GMM components are analyzed. The main contribution of this paper is to provide one possible structure to segment and cluster the path primitives and train the MPs within each type of path primitives. Besides, the results also indicate the suitable parameters to train the MPs for path-tracking applications.

\par The remains of this paper are organized as follows. Section \uppercase\expandafter{\romannumeral3} presents the GMM and GMR methods that we use to solve the problem. Section \uppercase\expandafter{\romannumeral4} introduces the data collection system and model training process. Section \uppercase\expandafter{\romannumeral5} shows the experiment results and discussion. Finally, the conclusions are given in Section \uppercase\expandafter{\romannumeral6}.

\section{Problem Statement}

Learning motion primitives is to extract basic elementary vehicle paths and driver operations from human demonstrations. After the motion primitives have been learned, the lateral operation commands are then generalized according to the known parameters of the motion primitives. 

\par The driving data is used as a basis for learning of path-tracking skills. The following variables are defined to represent the vehicle path points, driver lateral operation commands and motion primitives.

\begin{itemize}		

\item \begin{math}	{\boldsymbol{O}_t}={[\Delta {\theta _t},{v_t}]^{\rm{T}}} \in {\mathbb{R}^{2 \times 1}} \end{math} is the definition of the vehicle path point at time $t$, where \begin{math}  \Delta {\theta _t} = {\theta _t} - {\theta _{t - 1}} \end{math} is the course deviation between the time $t$ and $t-1$, and ${v_t}$ is the velocity of the path point.

\item \begin{math}  {\delta _t} \in {\mathbb{R}^{1 \times 1}} \end{math} is the state of the steering wheel angle at time $t$.

\item \begin{math}  {\boldsymbol{P}_t} = {[\{ {\boldsymbol{O}_{t - n\Delta t}},{\delta _{t - n\Delta t}}\}...\{ {\boldsymbol{O}_{t + n\Delta t}},{\delta _{t + n\Delta t}}\} ]^{\rm{T}}} \in {\mathbb{R}^{3 \times (2n + 1)}}   \end{math} is the definition of motion primitive at time $t$, where $n \in {{\rm{N}}^*}$, $ \Delta t = 0.1{\rm{s}} $.

\end{itemize}	

\par Therefore, the general form of the proposed motion primitive application for path tracking is presented as
\begin{gather}
{\lambda _i}({\boldsymbol{P}_t}):{\boldsymbol{P}_{{t_0}}}\_\boldsymbol{in} \mapsto {\boldsymbol{\delta} _{{t_0}:{t_0} + n\Delta t}} \label{equ1}\\
{\boldsymbol{P}_{{t_0}}}\_\boldsymbol{in} = [\{ {\boldsymbol{O}_{{t_0} - n\Delta t}},{\delta _{{t_0} - n\Delta t}}\} ,...,\{ {\boldsymbol{O}_{{t_o}}},{\delta _{{t_0}}}\}]\label{equ2}
\end{gather}
\par The Eq.\ref{equ1} is to generate the predicted steering wheel angle sequences ${\boldsymbol{\delta} _{{t_0}:{t_0} + n\Delta t}}$ with the motion primitive input variables ${\boldsymbol{P}_{{t_0}}}\_\boldsymbol{in}$ which can be observed at time $t_0$ corresponding to the trained model ${\lambda _i}$ classified by the path cluster labels.

\section{Learning and Generalizing Motion Primitives}

\par In our upper-level framework, the path primitives are defined by the cluster labels. The path primitives can be segmented and represented by a set of features, whereby the features are computed from the raw data of path point sequences. Finding an appropriate feature set to segment and represent the path primitive is essential for the clustering algorithm. One suitable way to solve this problem is to use the zero crossing course deviation as an intuitive criterion to obtain the segmentation of paths, and apply GMM to cluster the selected features of the segmented path. 

\par The switch between different path primitives is considered to be discrete. Therefore, only one path primitive is active at a time. The motion primitives are learned and generalized based on the chosen path primitive in our lower-level framework. Three kinds of features are selected to train the motion primitive models: the previous, current and future state of path point sequences and steering wheel angle sequences. The motion primitives are trained by using GMM and generalized by applying GMR.

\subsection{Segmenting and Representing Path Primitives} 

\par Given a path point sequence data, path primitive segments are generated through the following steps: Segment labeling is of thresholding on the course deviation data, and the data are segmented into left, right and neutral. Although the segment labeling is neutral, fluctuations exist in the course deviation data. So with this processing, the lane keeping behavior and the changing of direction are served as the segment point of the path point sequence.

\par Due to the limited number of path samples and noisy sensor data, it is crucial to select the most significant features to avoid the over-fitting problem of path primitive model training. As a result, four dominant features are defined and listed in Eq \ref{equ3}. The selected features address the aspects of steering operation time, course angle characteristics and velocity characteristics. 

\par In the situation of path primitive model training, the feature dataset ${\boldsymbol{x}_{path}}(i)$ for each path segment is defined as
\begin{gather}
{\boldsymbol{x}_{path}}(i) = [td(i),ave\_cd(i),max\_cd(i),ave\_vel(i)] \label{equ3}
\end{gather}

\noindent where $td(i)$ is the time duration of the segmented path primitive $i$, $ave\_cd(i)$, $max\_cd(i)$ and $ave\_vel(i)$ are the average course deviation, maximum course deviation and average velocity of the segmented path primitive $i$ separately. 

\par The GMM is applied to train the cluster model both for the path primitives and motion primitives. The GMM is represented as
\begin{gather}
G(\boldsymbol{x};\boldsymbol{\beta}) = \sum\limits_{i = 1}^K {{p_i}{g_i}(\boldsymbol{x};{\boldsymbol{\mu} _i},{\boldsymbol{\Sigma} _i})}  \label{equ4}\\
{g_i}(\boldsymbol{x};{\boldsymbol{\mu} _i},{\boldsymbol{\Sigma} _i}) = \frac{1}{{\sqrt {{{(2\pi )}^d}\left| {{\boldsymbol{\Sigma} _i}} \right|} }}{e^{ - \frac{1}{2}{{(x - {\boldsymbol{\mu} _i})}^T}{\boldsymbol{\Sigma} _i}^{ - 1}(x - {\boldsymbol{\mu} _i})}} \label{equ5}
\end{gather}

\noindent where $\boldsymbol{x}$ is a set of d-dimension sequence, $x = \left\{ {{\boldsymbol{x}_i}} \right\}_{i = 1}^n$ with ${x_i} \in {\mathbb{R}^{d \times 1}}$, ${\mu _i} \in {\mathbb{R}^{d \times 1}}$ and ${\Sigma _i} \in {\mathbb{R}^{d \times d}}$ are mean vector and covariance vector of a single Gaussian Model, $p_i$ is the weight coefficient of a single Gaussian Model and $\sum\nolimits_{i = 1}^n {{p_i} = 1}$, $\boldsymbol{\beta}  = \left\{ {{\boldsymbol{\beta} _i}} \right\}_{i = 1}^n$ with ${\boldsymbol{\beta} _i} = \left\{ {{\boldsymbol{\mu} _i},{\boldsymbol{\Sigma} _i},{p_i}} \right\}$

\par Maximum Likelihood Estimation of the model parameters is achieved iteratively using the Expectation-Maximization (EM) algorithm. To avoid getting trapped into poor local minima, a k-means clustering technique is applied to solve the initial estimation problem. Besides, the number of the mixture model k is determined by the Bayes Information Criterion (BIC).

\par After the model has been trained by utilizing the selected features, the path point sequences can be labeled by the trained model. The cluster label of path segment is represented as
\begin{gather}
labe{l_{path}}(i) = \arg \mathop {{\rm{max}}}\limits_{1 \le j \le k} \{ Pr(j|{\boldsymbol{x}_{path}}(i))\}   \label{equ6}\\
Pr(j|{\boldsymbol{x}_{path}}(i)) = \frac{{{p_j}{g_j}({\boldsymbol{x}_{path}}(i);{\boldsymbol{\mu} _i},{\boldsymbol{\Sigma} _i})}}{{\sum\nolimits_{l = 1}^k {{p_l}{g_l}({\boldsymbol{x}_{path}}(i);{\boldsymbol{\mu} _l},{\boldsymbol{\Sigma} _l})} }} \label{equ7}
\end{gather}

\noindent where $Pr(j|{\boldsymbol{x}_{path}}(i))$ is the probability of data ${\boldsymbol{x}_{path}}(i)$ being in Gaussian model $j$.

\par Although the motion primitives which will be discussed in the next section have already taken previous states of the path points into consideration, the time scale is limited in order to restrict the dimensions of GMM. Due to the requirement of long timescale consideration without increasing the dimensions of GMM, the regrouped training data of motion primitives are based on the previous, current and future cluster labels of path primitives. The path primitive $Path(i)$ is defined as
\begin{gather}
Path(i) = \left\{ {labe{l_{path}}(i - 1),labe{l_{path}}(i),labe{l_{path}}(i + 1)} \right\}  \label{equ8}
\end{gather}

\subsection{Learning Motion Primitives}
\par In our lower-level framework, the motion primitives are learned based on the type of path primitive. Therefore, the collected driving data are divided into different groups which are then used as the training dataset for different kinds of motion primitive models. The number of motion primitive models is equal to the number of path primitive type.

\par The output grouped training data for motion primitive model $\lambda _i$ is shown as follows:
\begin{gather}
{\xi _i} = [\Delta {\theta _i},{v_i},{\delta _i},ph{t_i}] = [{\boldsymbol{D}_i},ph{t_i}] \label{equ9}\\
\mathop  \cup \limits_{pht = 1} {\boldsymbol{D}_{pht}} = \boldsymbol{D} {\rm{and}}\mathop  \cap \limits_{pht = 1} {\boldsymbol{D}_{pht}} = \emptyset ,pht = 1,2...{n_1} \label{equ10}
\end{gather}

\noindent where $pht$ represents the path primitive type, $\boldsymbol{D}$ is the total collected driving data, $\boldsymbol{D}_{pht}$ is the regrouped dataset according to the different path primitive type, $n_1$ is the number of path primitive type.

\par After the data has been regrouped, the GMMs are applied to train the motion primitive models respectively. The single training dataset       ${\boldsymbol{x}_{m}}({t_i})$ is defined as
\begin{gather}
{\boldsymbol{x}_{m}}({t_i}) = [\boldsymbol{P}_p,\boldsymbol{P}_c,\boldsymbol{P}_f]\\
\boldsymbol{P}_p = [{\boldsymbol{O}_{{t_i} - {n_2}\Delta t:{t_i} - \Delta t}},{\boldsymbol{\delta} _{{t_i} - {n_2}\Delta t:{t_i} - \Delta t}}]\label{eq_n2}\\
\boldsymbol{P}_c = [{\boldsymbol{O}_{{t_i}}},{\boldsymbol{\delta} _{{t_i}}}]\\
\boldsymbol{P}_f = [{\boldsymbol{O}_{{t_i} + \Delta t:{t_i} + {n_3}\Delta t}},{\boldsymbol{\delta} _{{t_i} + \Delta t:{t_i} + {n_3}\Delta t}}]\label{eq_n3}
\end{gather}

\noindent where ${n_2},{n_3} \in {{\mathbb{N}}^*}$ are the value of time scale that takes into consideration when training the motion primitive model.

\par The model training process is the same as path primitive model training. When the model has been trained, the parameters of each motion primitive GMMs could be calculated as 

\begin{gather}
{\boldsymbol{\beta} _{pht}}_{_i} = \left\{ {{\boldsymbol{\mu} _{pht}}_{_i},{\boldsymbol{\Sigma} _{pht}}_{_i},{p_{pht}}_{_i}} \right\} \label{equ12}
\end{gather}

\subsection{Generalizing Motion Primitives}
To reconstruct a general form for the future steering wheel angle we apply GMR. The motion primitive variables ${\boldsymbol{P}_{{t_0}}}\_\boldsymbol{in}$ at time $t_0$ are used as the input states of the motion primitive model. And the corresponding values ${\boldsymbol{\delta} _{{t_0}:{t_0} + n\Delta t}}$ are estimated through regression. 

\par For each type of motion primitive GMM models, the input and output parameters are separated. The mean and covariance matrix of the Gaussian component $k$ are defined as:
\begin{gather}
{\boldsymbol{\mu} _k} = \left\{ {{\boldsymbol{\mu} _{in.k}},{\boldsymbol{\mu} _{es,k}}} \right\} \label{equ13}\\
{\boldsymbol{\Sigma} _k} =
\begin{bmatrix}
{\boldsymbol{\Sigma} _{in,k}} & {\boldsymbol{\Sigma} _{ines,k}} \\
{\boldsymbol{\Sigma} _{esin,k}} & {\boldsymbol{\Sigma} _{es,k}}
\end{bmatrix}
\label{equ14}
\end{gather}

\par The conditional expectation and estimated conditional covariance of ${\boldsymbol{\delta} _{{t_0}:{t_0} + n\Delta t}}$ with the input of ${\boldsymbol{P}_{{t_0}}}\_\boldsymbol{in}$ are defined as:
\begin{gather}
{\hat{\boldsymbol{\mu}} _{es,k}} = {\boldsymbol{\mu} _{es,k}} + {\boldsymbol{\Sigma} _{esin,k}}{({\boldsymbol{\Sigma} _{in,k}})^{ - 1}}({\boldsymbol{P}_{{t_0}}}\_\boldsymbol{in} - {\boldsymbol{\mu} _{in.k}})  \label{equ15}\\
{\hat{\boldsymbol{\Sigma}} _{es,k}} = {\boldsymbol{\Sigma} _{es,k}} - {\boldsymbol{\Sigma} _{esin,k}}{({\boldsymbol{\Sigma} _{in,k}})^{ - 1}}{\boldsymbol{\Sigma} _{ines,k}} \label{equ16}
\end{gather}

\noindent where ${\hat{\boldsymbol{\mu}} _{es,k}}$ and ${\hat{\boldsymbol{\Sigma}} _{es,k}}$ are mixed based on the probability that corresponds with ${\boldsymbol{P}_{{t_0}}}\_\boldsymbol{in}$ in Gaussian component $k$. 
\begin{gather}
{\beta _k} = \frac{{p({\boldsymbol{P}_{{t_0}}}\_\boldsymbol{in}|k)}}{{\sum\nolimits_{i = 1}^K {p({\boldsymbol{P}_{{t_0}}}\_\boldsymbol{in}|i)} }} \label{equ17}
\end{gather}

\par The final estimation of ${\boldsymbol{\mu}} _{es}$ and $\boldsymbol{\Sigma} _{es}$ are given by a mixture of $K$ components using Eq. \ref{equ15}, Eq. \ref{equ16} and Eq. \ref{equ17}.
\begin{gather}
{\hat{\boldsymbol{\mu}} _{es}} = \sum\limits_{k = 1}^K {{\beta _k}{{\hat{\boldsymbol{\mu}}}_{es,k}}}   \label{equ18}\\
{\hat{\boldsymbol{\Sigma}} _{es}} = \sum\limits_{k = 1}^K {{\beta _k}^2} {\hat{\boldsymbol{\Sigma}} _{es,k}}  \label{equ19}
\end{gather}

\par Therefore, by predicting $\left\{ {{{\hat{ \boldsymbol{\mu}} }_{es}},{{\hat{\boldsymbol{\Sigma}}}_{es}}} \right\}$ at different time $t_0$, a generalized form of the motion primitive ${\hat{ \boldsymbol{P}} _t} = \left\{ {{\boldsymbol{P}_{{t_0}}}\_\boldsymbol{in},{{\hat{ \boldsymbol{\delta}} }_{{t_0}:{t_0} + n\Delta t}}} \right\}$ and the associated covariance matrix ${\hat{\boldsymbol{\Sigma}} _{\delta \_es}}$ are produced.

\section{Data Collection and Model Training}

In this section, we will introduce our data collection system and present an overview of our dataset. Besides, the details of data training process which includes the preprocessing and different model training variables are presented. Finally, the performance index of the prediction accuracy is discussed.

\subsection{Data Collection}

The BIT intelligent vehicle platform was used to collect the driving data (see in Fig. \ref{fig１(a)}). Nine drivers of different ages and personalities have participated in the driving data collection experiments in Beijing. The driving situations contain ring roads, highways, intersections, etc. The driving operations include lane keeping, lane changing, overtaking and other typical driving operations.

\subsection{Data Training Process}

\subsubsection{Preprocessing}

The emergency driving situations are removed from the training dataset manually with the help of collected camera data. Moving average filter with a window size W=5 is utilized to smooth the entire extracted database.

\subsubsection{Motion Primitive Model Training Process}

The different training process or different types of motion primitive models can be concluded by four parameters: the number of different path primitive $n_1$, the value of previous time scale $n_2$, the value of prediction time scale $n_3$ and the number of GMM components $n_4$. In order to evaluate the influence of the components number on motion primitive model, the three parameters are selected as follows.

\begin{itemize}	

\item ${n_1} \in \{ {1,3,27} \}$ are selected to evaluate the influence of the number of path primitives. When $n_1$ is set to 1, there is no upper-level. The motion primitives are directly learned from the driving data without regrouping. When $n_1$ is set to 3 which is determined by the BIC, the upper-level is activated. When $n_1$ is set to 27, the timing relationship of path clustering labels is considered (see in Eq.\ref{equ8} ). 

\item ${n_2} \in \{ { - 1,0,1,2,3} \}$ are selected to evaluate the effect of previous time scale. When $n_2$ is set to -1, even the current state of the steering wheel angle is not considered during the model training process (see in Eq. \ref{eq_n2}).

\item The prediction time scale $n_3$ is always set to 50. So the prediction time domain is selected as 5s no matter how the training process changes. Therefore, the evaluation indicators could be unified (see in Eq. \ref{eq_n3}).

\item ${n_4} \in \{ {3,6,9}\}$ are selected to evaluate the influence of the number of GMM components that utilized in the training process of motion primitive model.

\end{itemize}

\section{Results}

\subsection{Segmenting and Clustering Results of Path Primitives}
For all the driving data, our segmentation algorithm was always able to find the correct segmented point. By utilizing the feature dataset of each path primitive, the cluster model for path primitives was trained. The mean matrixes of each Gaussian component were shown in Table \ref{table_1}.
\begin{table}
 \centering
 \begin{tabular}{c c c c c}
  \toprule
  Label & $td$(s) & $ave\_{cd}$($^\circ$) & $max\_{cd}$($^\circ$) & $ave\_{vel}$(km/h)\\ 
  \midrule
  1 & 8.29 & 0.15 & 0.45 & 33.45 \\
  2 & 0.24 & 0.008 & 0.01 & 50.31\\
  3 & 1.15 & 0.024 & 0.05 & 54.69 \\
  \bottomrule
 \end{tabular}
 \caption{Clustering results of the three path primitives based on GMM.}
 \label{table_1}
\end{table}
\begin{table}
 \centering
 \begin{tabular}{c c c c c c}
  \toprule
  $type$ & $n_1$ & $n_2$ & $n_4$ & $ave\_err$($^\circ$) & $var$\\ 
  \midrule
  1 & 1 & 0 & 3 & 2.12 & 20.72 \\
  1 & 3 & 0 & 3 & 1.91 & 18.28 \\
  1 & 27 & 0 & 3 & 1.87 & 15.84 \\
  2 & 27 & -1 & 6 & 2.84 & 42.61 \\
  2 & 27 & 0 & 6 & 1.54 & 9.84 \\
  2 & 27 & 1 & 6 & 1.44 & 7.43 \\
  2 & 27 & 2 & 6 & 1.41 & 8.2 \\
  2 & 27 & 3 & 6 & 1.28 & 5.14 \\
  3 & 27 & 1 & 3 & 1.43 & 7.92 \\
  3 & 27 & 1 & 6 & 1.44 & 7.43 \\
  3 & 27 & 1 & 9 & 1.17 & 6.39 \\
  3 & 3 & 1 & 3 & 1.91 & 18.28 \\
  3 & 3 & 1 & 6 & 1.88 & 12.93 \\
  3 & 3 & 1 & 9 & 1.82 & 10.89 \\
 \bottomrule
 \end{tabular}
 \caption{The table outlines the average errors $ave\_err$ of the predicted results compared to the original driving data and indicates the predicted uncertainty by using the variance of the estimation $var$. Experiment type 1-3 indicates the influence of upper-level results, previous time scales and different GMM component number  separately}
 \label{table_2}
\end{table}
\subsection{Predicting Results by Generalizing Motion Primitives}
The comparison results of different training parameters are shown in Table \ref{table_2}. The dataset of $path\_type = \{1,1,1\}$ was chosen to evaluate the model. Although the trend of the comparison results of each path primitive type was just the same, the selected path primitive type was the most obvious. 

\par
\begin{figure}
\centering
\subfloat[Low-speed with sharp steering]{
\begin{minipage}[t]{0.45\textwidth}
\centering
\includegraphics[scale=0.315]{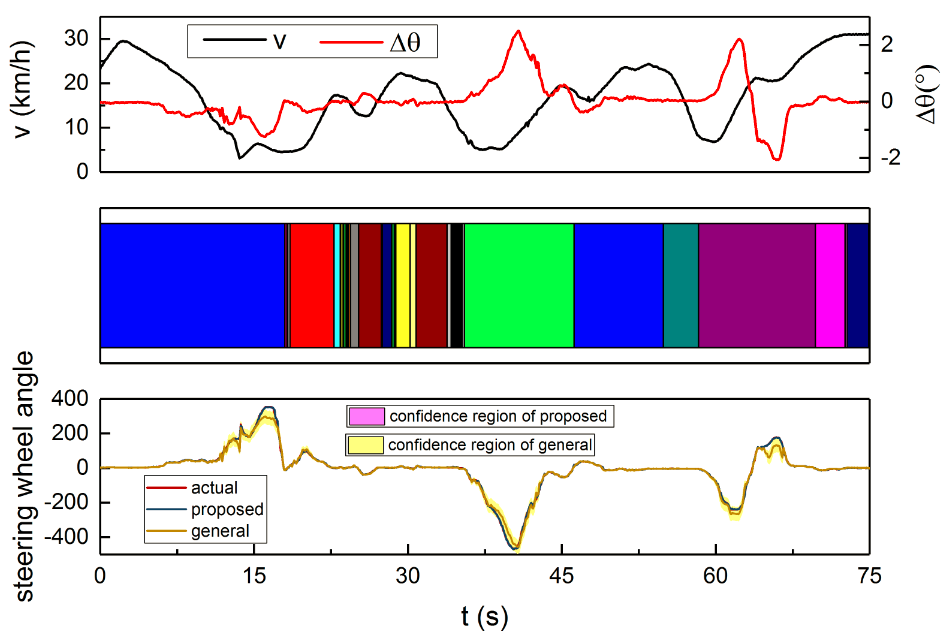} 
\end{minipage}
\label{fig2(a)}
}

\subfloat[Medium-speed with general steering]{
\begin{minipage}[t]{0.45\textwidth}
\centering
\includegraphics[scale=0.315]{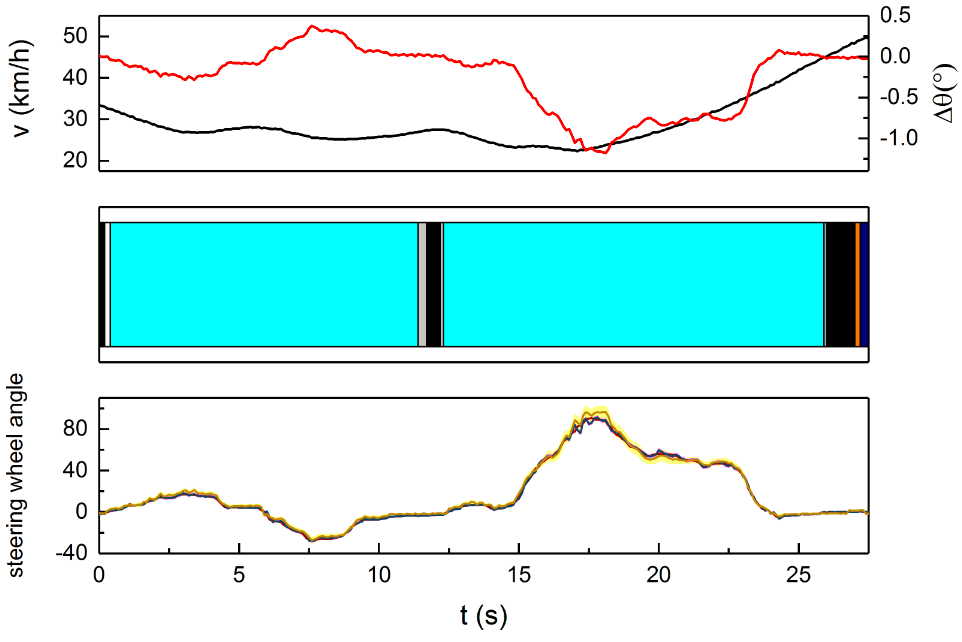} 
\end{minipage}
\label{fig2(b)}
}

\subfloat[High-speed with direction correction]{
\begin{minipage}[t]{0.45\textwidth}
\centering
\includegraphics[scale=0.315]{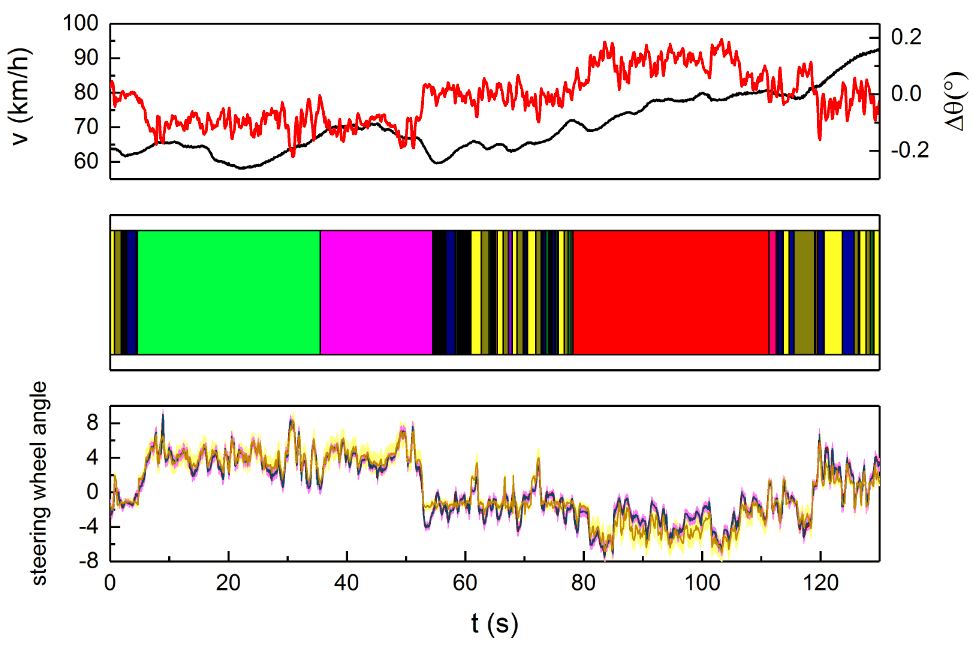}
\end{minipage}
\label{fig2(c)}
}

\subfloat[High-speed with straight line]{
\begin{minipage}[t]{0.45\textwidth}
\centering
\includegraphics[scale=0.315]{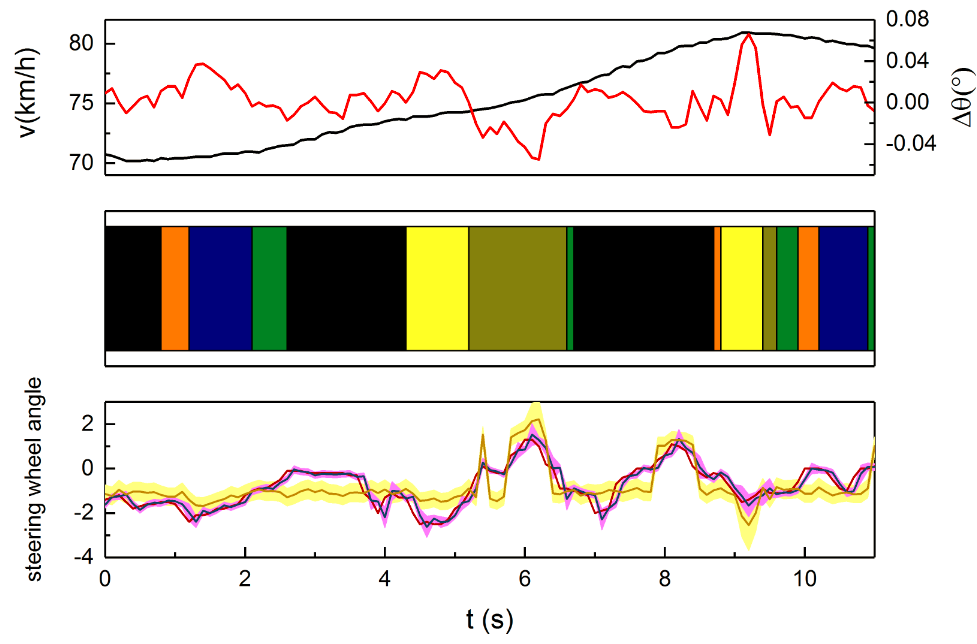} 
\end{minipage}
\label{fig2(d)}
}
\caption{The prediction results of four typical driving situations. In each subgraph, the desired path information, path type identification result and prediction result are presented in order. The shaded region corresponds to the standard deviation. Two methods were compared here, one is called the proposed method(two-level structure approach) with a training parameters$\{n_1=27,n_2=1.n_4=3\}$, the other one is called the general method(the reference model) with a training parameters $\{n_1=1,n_2=-1.n_4=9\}$ }
\label{fig2}
\end{figure}

 Finally, four typical driving situations were chosen to present and compare the prediction results of our methods. (see Fig. \ref{fig2}). Different colors within each segmentation result illustrated different types of MPs, and the colors in the four segmentation results are linked with each other. The general GMM-GMR model without the upper-level was chosen as the reference to highlight the advantages of our two-layer structure and the importance of taking the current state $\boldsymbol{P}_c$ into consideration.

\subsection{Discussion}

The benefit of our approach becomes quite obvious when comparing the results of the two-level structure approach with the referenced general GMM-GMR model, shown in Fig. \ref{fig2}. Despite achieving comparable predicted results, the prediction of the reference model works worse in generalizing MPs. From the driving situation a to situation d, the average errors of our approach is $5.26^\circ,1.03^\circ,0.54^\circ,0.41^\circ$ lower than the reference, and the standard deviation is $15.16^\circ,1.59^\circ,0.36^\circ,0.21^\circ$ lower than the reference.

\par The presented results in Table \ref{table_2} (type=1) also emphasize the advantage of the two-level training structure over direct training process. The path primitives were able to regroup the training data, thus increasing the predicted accuracy by $9.91\%$. The reason is the great difference between the path primitives, especially in the aspect of time duration and course deviation which are represented by Gaussian components in Table \ref{table_1}. So it is hard to train a signal MP model which can handle all the driving situations. However, our approach to segment the path primitives uses intuitive criterion. And there are some frequent switching of the path types, it is extremely obvious in Fig \ref{fig2(c)}. Actually some of the segmented points can be merged.

\par Besides, the previous time scale which has been taken into consideration in the MPs is another crucial aspect of the training process. If we do not consider the current state in model training, the average errors increase by $45.77\%$ and the average variance increase by $76.91\%$ (see Table \ref{table_2} (type=2)). The reason is that the lateral operation command is a continuous variable. So it is quite important to introduce some previous reference points in the MPs. However, the improvement in the prediction remains essentially unchanged when the past time scales become larger than 1.

\par Finally, the number of GMM components also influences the predicting accuracy, but the improvement is not as significant as the hierarchical structure and time scales (see Table \ref{table_2}). Considering the predicting accuracy and computational complexity, the overall winner is the training parameter set $\{n_1=27,n_2=1.n_4=3\}$. 

\section{Conclusion and Future Work}
In this paper, we proposed to use a two-level structure for learning and generalizing motion primitives. Based on this, path primitives were segmented first and then clustered by applying GMM. We presented how the observed path data can be incorporated into the GMM representation and what kind of typical features have been chosen to identify each path primitive. This leads to the regrouping of driving data based on the type of path primitives which is the main benefit of our structure. We evaluated different time scales and GMM components when applying GMM to train the MPs model and utilizing GMR to generalize the model. Our approach was validated with the collected driving data, one typical path primitive type was chosen to discuss the influence of training parameters and four typical driving situations were chosen to evaluate the predicting accuracy. We found that our approach is able to learn and generalize the MPs from driving data by using the suitable training parameters.

\par In future work, we aim at learning MPs considering the road conditions, combining the proposed approach with some stable path tracking methods and optimizing the MPs model online.

\bibliographystyle{unsrt}
\bibliography{root}

\end{document}